\documentclass[report]{ieeeconf}  
\IEEEoverridecommandlockouts                              
\usepackage{paralist}
\usepackage{graphics} 
\usepackage{mathptmx} 
\usepackage{times} 
\usepackage{amsmath} 
\usepackage{amssymb}  
\usepackage{graphicx} 
\usepackage{algorithm,algpseudocodex}
\usepackage{xargs}
\usepackage[colorinlistoftodos,prependcaption,textsize=tiny, textwidth=1.9cm]{todonotes}
\usetikzlibrary{mindmap,trees,fit,calc,arrows,backgrounds,decorations}
\usetikzlibrary{decorations.pathmorphing,shadows,positioning,shapes,snakes,patterns,automata}
\usetikzlibrary{shapes.multipart}

\newcommand{\reals}{\mathbb{R}}

\newcommand{\timestep}{h}

\newcommand{\outsig}{y}

\newcommand{\neuralnet}{\mathcal{N}}

\newcommand{\traj}{\gamma}
\newcommand{\data}{d}
\newcommand{\dataset}{\mathcal{D}}

\newcommand{\F}{\Diamond}

\newcommand{\globally}{\Box}
\newcommand{\G}{\Box}

\newcommand{\until}{\mathcal{U}}
\newcommand{\spec}{\varphi}
\newcommand{\true}{\top}

\newcommandx{\alexnote}[2][1=]{\todo[author=Alex,backgroundcolor=white,linecolor=blue,bordercolor=blue,#1]{#2}}
\newcommandx{\ihnote}[2][1=]{\todo[author=Inzemam,backgroundcolor=white,linecolor=red,bordercolor=red,#1]{#2}}
\newcommandx{\nknote}[2][1=]{\todo[author=Nikos,backgroundcolor=white,linecolor=magenta,bordercolor=red,#1]{#2}}
\newcommandx{\thaonote}[2][1=]{\todo[author=thao,backgroundcolor=white,linecolor=magenta,bordercolor=blue,#1]{#2}}
\newcommandx{\hidenote}[2][1=]{\todo[disable,#1]{#2}}
\newtheorem{problem}{Problem}
\newtheorem{definition}{Definition}

\def \tPara {\tau}
\def \sPara {s}

\def \similarity {\sigma}
\def \Plant {S}

\newcommand{\Status}{\ensuremath{\mathit{Status}}}
\newcommand{\NSamples}{\ensuremath{\mathit{NeighSamples}}}
\newcommand{\InitSamples}{\ensuremath{\mathit{InitSamples}}}
\newcommand{\InitTraces}{\ensuremath{\mathit{InitTraces}}}
\newcommand{\CexTraces}{\ensuremath{\mathit{CexTraces}}}
\newcommand{\FixedTraces}{\ensuremath{\mathit{FixedTraces}}}
\newcommand{\Cover}{\ensuremath{\mathit{Cover}}}

\usepackage{fancyhdr}
\title{\LARGE \bf
Counter-Example Guided Imitation Learning of Feedback Controllers from Temporal Logic Specifications}
\author{
Thao Dang$^{1}$, Alexandre Donz{\'e}$^{2}$, Inzemamul Haque$^{3}$, Nikolaos Kekatos$^{4}$, Indranil Saha$^{3}$  
\thanks{$^*$ This work is partially supported by the Indo-French Collaborative 
research project FOVERAS funded by IFCPAR/CEFIPRA, joint 
French-Japanese ANR-JST project CyPhAI, and the Auvergne-Rh{\^o}ne-Alpes Region Project DetAI.}
\thanks{$^{1}$ Univ. Grenoble Alpes, CNRS, Grenoble INP, VERIMAG, Grenoble, France, 
          {\tt\small thao.dang@univ-grenoble-alpes.fr}}%
\thanks{$^{2}$ Decyphir SAS, Moirans, France,
        {\tt\small alex@decyphir.com  }}%
\thanks{$^{3}$ Department of Computer Science and Engineering, IIT Kanpur, India,
        {\tt\small \{inzemam, isaha\}@cse.iitk.ac.in }}%
\thanks{$^{4}$ Aristotle University of Thessaloniki, Thessaloniki, Greece,   {\tt\small nkekatos@csd.auth.gr}
          {\tt\small  }}%
  }        

\begin{document}

\begin{titlepage}

\begin{figure}
\begin{minipage}[h]{.15\linewidth}
\centering
\includegraphics[height=2cm]{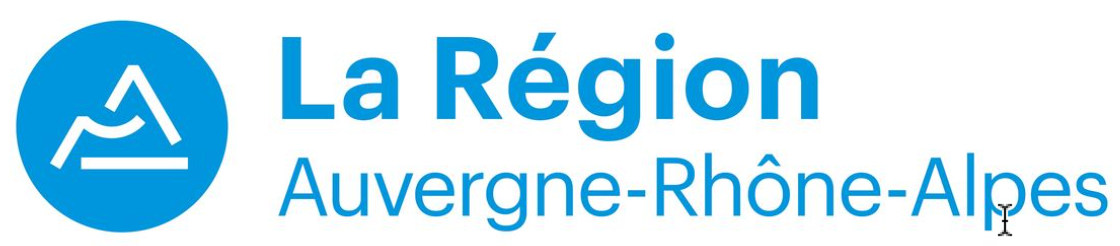}
\end{minipage} \hfill
\end{figure}

\maketitle

\end{titlepage}

\maketitle


\begin{abstract}
    We present a novel method for imitation learning for control requirements expressed using Signal Temporal Logic (STL). More concretely we focus on the problem of training a neural network to imitate a complex controller. The learning process is guided by efficient data aggregation based on counter-examples and a coverage measure. Moreover, we introduce a method to evaluate the performance of the learned controller via parameterization and parameter estimation of the STL requirements. We demonstrate our approach with a flying robot case study.  
\end{abstract}

\section{Introduction}
The aim of this work is to integrate formal specification and validation techniques in the Imitation Learning (IL) methodology for synthesizing feedback controllers for complex dynamical systems. While formal methods have the advantage of rigorous formalization and reasoning, they are very limited in the complexity and scalability of the problems that can be practically solved. Imitation Learning, also known as learning from demonstrations, involves the process of learning how to mimic the behavior of an expert by observing their actions in a given task~\cite{NIPS1996_68d13cf2}. It has many successful applications in various fields such as robotics, natural language processing, image and speech recognition.

In this work, we focus on the problem of training a neural network (NN) (playing the role of a learner) to imitate a complex controller (playing the role of an expert). The ultimate goal is to replace this complex controller with a trained NN. NNs have long been used to control dynamical systems from inverted pendulums to quadcopters, 
learning from scratch to control the plant by maximizing an expected reward, e.g.~\cite{hagan2002introduction,Nicol08}. 
NNs can also be trained to replace an existing controller that is unsatisfactory for non-functional reasons, e.g., computationally expensive (consider model-predictive control~\cite{mpc}), slow, or energy intensive. 
A well-trained NN controller can provide similar control performance much faster and is readily implemented on cheap and energy-efficient embedded platforms~\cite{varshney2019deepcontrol}.



To make such an imitation learning framework more formal and more efficient, we add the following novel features: 
(i) a formalization of performance evaluation for both the learner's and expert's policies using their abilities to satisfy requirements specified by temporal logic, 
(ii) a leverage of the power of existing temporal logic property falsification tools to create training data that matter, (iii) a new method of data aggregation in order to guarantee a good performance of NN in terms of imitation and generalization.


To explain these features, let us first point out some major difficulties in this problem. Training a NN to imitate a feedback controller is more complex than the problem of approximating a function using pairs of input and output values, since feedback controllers can themselves be stateful dynamical systems. We identify the following difficulties in data generation by executing the nominal controller in closed loop:
\begin{itemize}
    \item 
    \emph{Infinite behavior space.} 
    The behavior space is not only large but can also be infinite. It is thus important to define a coverage measure to quantify how representative the generated training data is.
    \item \emph{Non-uniform accuracy.} Depending on the control requirement, the NN may need to be very precise around some region of the state space while in other regions a rough approximation is acceptable. 
\end{itemize}
%
%
%
The problem of non-uniform accuracy is particularly pronounced when the requirement depends on time or sequences of events. This is frequently the case in control applications, where properties such as \emph{rise time}, \emph{settling time}, and \emph{overshoot} are typical. We consider complex properties that can include not only time but also causal relationships. They can be described in \emph{Signal Temporal Logic} (STL), a formal language that finds widespread use in formal methods and increasing adoption in industry \cite{bartocci2018specification}.

While observing closed-loop behaviors may reduce the number of  behaviors to be sampled, we still need to find good training samples  
%
 that are relevant to an STL property. To do this, we find counter-examples that are closed-loop behaviors violating this property by leveraging the existing falsification tools~\cite{breach}. A falsification process can also be useful in providing correctness guarantees for the resulting NN. Indeed, if no counter-example is found after a sufficiently large number of scenarios, we consider the NN controller satisfactory and stop. If a counter-example is found, we replay the nominal controller from the counter-example situation in order to obtain new training data, and retrain the neural network. This new data creation is crucial for the efficiency of the process of correcting counter-examples as well as assuring good generalization of the NN. 

{\bf Related Work.} 
Several comprehensive surveys exist 
 \cite{osa2018algorithmic} on imitation learning. A common approach is behavioral cloning \cite{michie1990cognitive} for which the main problem is \emph{compounding error}, when the assumption of independent and identically distributed data between training and testing data is not valid for sequential predictions. A common solution to this problem is  Dataset Aggregation (DAgger) which was proposed in~\cite{ross2011reduction}. It is an iterative algorithm which improves on behavioral cloning by training on a dataset that better resembles the observations the trained policy is likely to encounter. In this paper, we propose a similar approach in the sense that the expert is queried for good actions and new data are created and aggregated in the dataset. However, our goal is to design an approach that learns the dynamics of \emph{desirable} behavior and not the dynamics of \emph{possible} behavior.  Applications of imitation learning for Model Predictive Control have been proposed in~\cite{acerbo2021safe}.  A recent approach to "compress" an MPC into a NN using robust tube MPC is proposed in~\cite{tagliabue2022efficient}. In~\cite{chow2023parallelized}, the need to replace an MPC controller with a NN is exemplified for planning purposes. 
Our approach is different from the previous works as we use counter-example guided synthesis and a combination of coverage and PSTL formal specifications.  There are counter-example-based approaches similar to ours though - for instance, some use counter-example exploration (adversarial sample) to train NNs that seek to satisfy a given property  expressed  in temporal logic (see~\cite{yaghoubi2019worst}) or through a reference trajectory (see~\cite{ClaviereDS19}). 
The closest to our work is \cite{ClaviereDS19}, where the behavior of an MPC is approximated with a NN to enable a robot to follow a reference trajectory. The NN is refined by generating additional training data from counter-examples. However, unlike our framework,  
their approach is limited to memoryless controllers, and they use tracking closeness as the sole criterion to identify counter-examples.  In our experience, a NN need not always closely track the nominal behavior to satisfy the specifications. It may have some characteristics that are better than the nominal controller, such as a smaller overshoot, or quicker stabilization in some areas of the state space. Such behaviors would be eliminated using their approach.

The rest of the paper is organized as follows. In Section~\ref{sec-problem}, we formalize the imitation learning problem. This requires definitions of control requirements specified using Parametric Signal Temporal Logic. We also propose a notion of policy performance to quantify the difference between the policy of the learner and the expert which is necessary to assess the imitation quality. Subsequently, in Section~\ref{sec-methodology} and Section~\ref{sec-aggregation}, we describe our dataset aggregation-based learning methodology. Finally, we demonstrate our approach on a robotic case study in Section~\ref{sec-casestudy}.

\section{Controller Imitation Learning Problem}
\label{sec-problem}
\def \C {\mathcal{C}}
\def \P {\mathcal{P}}

\def \taustab {\tau_{\text{st}}}
\def \tautrans{\tau_{\text{tr}}}
\def \piov{s_{\text{ov}}}
\def \pistab{s_{\text{st}}}
\def \mustab {\mu_{\text{st}}}
\def \fstab {\varphi_{\text{st}}}
\def \muov {\mu_{\text{ov}}}
\def\pmax{\overline{p}} 
\def\pmin{\underline{p}}

We consider a continuous-time plant $\Plant$ with state $x \in \reals^n$ that is controlled by an input signal $u$, observed through an output signal $y$ with dynamics\footnote{External disturbance can be included in the system dynamics, and the data generation process for learning can then be straightforwardly extended to cover the disturbance space.}:
$\dot{x}(t) =  f(x(t), u(t))$ and $y(t)= \zeta(x(t)).$ The control input $u$ is computed by a discrete-time controller $\mathcal{C}$ with state $z \in \reals^{n_z}$ and $z_{k+1} = f_c(z_k, y_k)$, $u_k = \upsilon(z_k, y_k)$, where $y_k$ is a discrete-time signal resulting from sampling the output $y(\cdot)$ with time step $\timestep$. The continuous-time control $u(\cdot)$ is a piece-wise constant function defined as $\forall t \in$ $[k\timestep, (k+1)\timestep]~ u(t) = u_k$ with $k = 0, 1, \ldots$. We denote the closed-loop system by $\mathcal{C}|| \Plant$. 


\subsection{Control Requirements}
We are given a nominal controller satisfying some requirement that captures essential qualitative properties while allowing some quantitative behavioral flexibility. This requirement is expressed using \emph{Parametric Signal Temporal Logic (PSTL)}, a formalism suitable for various control performance properties~\cite{bartocci2018specification}. 
We want to train a neural network-based controller achieving performance comparable to that of the nominal controller, as measured by \emph{valid} parameters of the PSTL requirement. 

\subsubsection{Signal Temporal Logic \cite{bartocci2018specification}}
An STL formula $\spec$ consists of atomic predicates along with logical and temporal operators. Atomic predicates are defined over signal values and have the form $g(\outsig(t))\sim 0$, where $g$ is a scalar-valued function over the signal $\outsig$ evaluated at time $t$ and $\sim \in \{ <,\leq, >, \geq, =, \neq\}$. 
Temporal operators ``always'' ($\globally$), ``eventually'' ($\F$), and ``until'' ($\until$) have the usual meaning and are scoped using intervals of the form $(a,b)$, $(a,b]$, $[a,b)$, $[a,b]$, or $(a,\infty)$, where 
$a,b\in \reals_0^{+}$ and $a<b$. If $I$ is a time interval, the following grammar defines the STL language.
\begin{equation}~\label{eqn:stl-gen}
\spec ~ := ~ \true \; | \; g(\outsig(t))\sim 0 \; | \; \neg \spec \; | \;
\spec_1 \wedge \spec_2 \; | \; \spec_1 \until_I \spec_2
\end{equation}
The $\F$ operator is formally defined as $\F_I \spec \triangleq \true \until_I \spec$, and the $\globally$ operator is defined as $\globally_I \spec \triangleq \neg (\F _I \neg \spec)$. When omitted, the interval $I$ is taken  to be $[0,\infty)$. 
 The ``always'' operator in $\globally\spec$ conveys that from the current time point onwards, $\varphi$ always holds. The ``eventually'' operator in $\F\spec$ means that there exists a time point in the future where $\spec$ holds. The ``until'' operator in $\spec_1\until\spec_2$ means that, starting from the current time point $\spec_1$ should hold continuously until a future time point where $\spec_2$ holds. Additionally, an interval $I$ can be combined with the operators to bound their scope to a segment of time in the future rather than the whole. For example, $\spec_1\until_{[a,b]}\spec_2$ holds true at time $t$ if $\spec_2$ holds true at some point $t'\in[t+a,t+b]$ and $\spec_1$ is always true in $[t,t']$. 
 
 
\subsubsection{Parametric Signal Temporal Logic~\cite{asarin_parametric_2011}}
Parametric STL (PSTL) is a variant of STL which makes it possible to replace numeric constants in an STL formula with symbolic variables or parameters. For instance, the formula $\varphi=\G_{[0, \tPara]} (\|y(t)\|<\sPara)$ with two parameters $\tPara$ and $\sPara$ expresses the requirement that during $\tPara$ seconds, the norm of signal $y$ should be less than $\sPara$. Formally, a PSTL formula is concretized into a STL formula by composing it with a valuation, defined as a mapping from symbolic parameters to reals. As an example, consider valuation $p: \{\tPara \rightarrow 2, \sPara \rightarrow 10\}$, then  $\varphi(p)$ is the STL formula $\varphi(p) = \G_{[0, 2]} (\|y(t)\|<10)$.\\

Given a PSTL formula $\varphi$ and a valuation $p$, a behavior $\traj$ satisfies $\varphi(p)$ is denoted by $\traj \models \varphi(p)$. By extension, we say that a valuation $p$ is satisfied by a controller $\mathcal{C}$ if for all behaviors $\traj$ of the closed-loop system $\mathcal{C}|| \Plant$, $\traj \models \varphi(p)$. 

The problem of finding a valuation $p$ such that $\mathcal{C}$ satisfies $\varphi(p)$ is sometimes called mining and can be seen (and solved) as a learning problem. In \cite{jin_mining_2015}, an approach is presented that works for a \emph{monotonic} PSTL formula. Intuitively a formula is monotonic if its  satisfaction is monotonic {\em w.r.t.} the valuation of each individual parameter. For example, $\varphi=\G_{[0, \tPara]} (\|y(t)\|<\sPara)$ is monotonic because if $\|y(t)\|$ is always smaller than $\sPara$ between time $0$ and $\tPara$, then clearly $\|y(t)\|$ is smaller than $\sPara'$ for any $\sPara'>\sPara$ and it is also smaller than $\sPara$ between time $0$ and $\tPara'$ for $\tPara'\leq \tPara$. Formally, the set $\P$ of all valuations is a subset of $\reals^{n_p}$, where $n_p$ is the number of parameters in the formula. Let $\preceq$ be the standard partial order on $\reals^{n_p}$, {\em i.e.}, $p=(p_1,\hdots, p_{n_p}) \preceq p'=(p'_1,\hdots, p'_{n_p})$ iff $\forall i \; p_i\leq p_i'$, then if $\forall \traj$ the Boolean function $p \rightarrow \traj \models \varphi(p)$ is monotonic in $p$, then the PSTL formula $\varphi$ is monotonic.
For illustration purposes, to characterize our controller performance, the PSTL formula $\Phi$ that we will use is as follows: 
\begin{eqnarray}
&\muov:=\|y(t)\|>\piov,\  \mustab:=\|y(t)\|<\pistab&  \label{eq:predicates}\\
&\fstab:= \neg \mustab  \Rightarrow  \F_{[0, \tautrans] } \G_{[0, \taustab]} \mustab& \label{eq:stab} \\ 
&\Phi:= \G \neg \muov \wedge \G \fstab& \label{eq:phi}
\end{eqnarray}

Equation~(\ref{eq:predicates}) defines the atomic predicates which check at a given time whether the signal norm is above $\piov$ (overshoot),
 or below $\pistab$ (defining a stabilization region around the equilibrium). Equation~\eqref{eq:stab} defines a formula requiring that if the system is not stabilizing ($\mustab$ not satisfied), then it should eventually stabilize, {\em i.e.}, after at most $\tautrans$ seconds, $\mustab$ should remain true for at least $\taustab$ seconds.
 




    
    
    
We can then see that $\Phi$ is monotonic. This follows from the monotonicity of atomic predicates and temporal operators $\G$ and $\F$ and the fact that each parameter appears only once in each sub-formula \cite{asarin_parametric_2011}. Monotonicity here is advantageous not only for the mining problem but also for computing a performance measure defined by Pareto fronts as discussed in the next section.

\subsection{Control Policy Performance Measure}
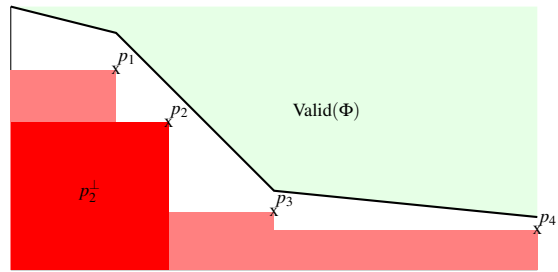
\begin{figure}
    \begin{center}
    \begin{tikzpicture}[scale=.7, transform shape]
      
      \draw[-] (0,0) -- (0,5) node[above] {};
      \draw[-] (0,0) -- (10,0) node[right] {};
      
      \fill[-,green!10] (0,5) -- (0,5) -- (10,5) -- (10,1) -- (5,1.5) -- (2,4.5) -- (0,5);
      \draw[-, thick] (10,1) -- (5,1.5) -- (2,4.5) -- (0,5);

      \def \down {
      \fill [-, red!50] (0, \y) -- (\x,\y) -- (\x, 0) -- (0,0); 
      \node at (\x,\y) {x} ;}

    \def \downleg {
      \node at (\x,\y) [] {$p_\k$};
      }
   \def \downplus {
      \fill [-, red!100] (0, \y) -- (\x,\y) -- (\x, 0) -- (0,0); 
      \node at (\x,\y) {x} ;}

      \def \up {
      \fill [-, green!100] (10,6) -- (\x , 6 ) -- (\x,\y) -- (10, \y) -- (10,6); 
      \fill (\x,\y) circle (2pt);}



    
      \def \x {2}  \def \y {3.8}     {\down}
      \def \x {2.2}  \def \y {4} \def \k {1}     {\downleg}
      
      \def \x {5}  \def \y {1.1}     {\down}
      \def \x {5.2}  \def \y {1.3} \def \k {3}     {\downleg}
      
      \def \x {10}  \def \y {0.75}     {\down}
      \def \x {10.2}  \def \y {.95} \def \k {4}     {\downleg}
 
      \def \x {3}  \def \y {2.8}     {\downplus}
       \def \x {3.2}  \def \y {3} \def \k {2}     {\downleg}
        \node at (1.5,1.5) [] {$p^{\bot}_2$};

      \node at (6,3) [] {$\text{Valid}(\Phi)$};
      
     

    \end{tikzpicture}
  \end{center}
    \caption{Volume estimation of False parameter set. Each $p_i$ outside the Valid set is close to the Pareto front and defines a closed hyper-box $p_i^\bot$ strictly included in False$(\Phi)$. Computing the volume of $\cup_i p_i^\bot$ yields an under-approximation of $vol(\text{False}(\Phi)$.}
    \label{fig:vol-approximation}
\end{figure}

In imitation learning, it is essential to have an appropriate measure of performance of control policies, especially when it is unclear what reward function is being optimized~\cite{DBLP:journals/corr/abs-1811-06711}. In our framework, we use the relation between the parameters in the PSTL requirements to compare the performance of different controllers. E.g., for the stabilization requirement~\eqref{eq:phi}, for a given size $\pistab$ of the neighborhood around the equilibrium, the smaller the stabilization time $\taustab$ is, the faster the controller is. 
Assume that  $\P(\Phi)$ (set of parameter valuations for $\Phi$) is compact. Any controller $\mathcal{C}$ defines a partition of this set into falsified and valid formulas: 
$$
\P(\Phi) = \text{False}(\mathcal{C}||\Plant, \Phi)\cup \text{Valid}(\mathcal{C}||\Plant,\Phi)
$$
 The False and Valid sets are separated by the set of {\em Pareto-efficient} parameter values, also called the {\em Pareto front} \cite{Pareto1912}; that means no parameter can be improved without compromising the others. We argue that the relative volumes of the False (and Valid) sets can be used to measure and compare performance of controllers. More specifically, we define the following measure of similarity:  
\begin{definition}[Control Policy Similarity]\label{def:similarity}
Given a plant $\Plant$ and two controllers $\mathcal{C}$ and $\mathcal{C}'$ designed to satisfy a PSTL requirement $\Phi$, the performance similarity between $\mathcal{C}$ and $\mathcal{C}'$ is defined as 
$\similarity_{\Plant,\Phi}(\mathcal{C}, \mathcal{C}')= \frac{vol(\text{False}(\mathcal{C'}||\Plant,\Phi)}{vol(\text{False}(\mathcal{C}||\Plant,\Phi)}$
where $vol$ is the volume of a set, assumed to be non-zero for $\text{False}(\mathcal{C}||\Plant,\Phi)$.  
\end{definition}
 Exact volume computation is difficult in general but monotonicity makes it easy to compute under-approximations. Indeed, consider a finite set of parameters $p_1, \hdots, p_k $ in False$(\mathcal{C}||\Plant, \Phi)$ and for each $p_i$, define $p^{\bot}_{i}$ the set of parameters dominated by $p_i$ according to $\preceq$. Intuitively, $p^{\bot}_{i}$ is a hyper-box with largest corner $p_i$. Then $\bigcup_i p^{\bot}_i \subset \text{False}(\mathcal{C}||\Plant, \Phi)$ and  $\sum_i vol(p^{\bot}_i) \leq vol(\text{False}(\mathcal{C}||\Plant, \Phi))$. Since $p^{\bot}_i$ is an hyperbox, its volume computation is trivial. This approximation is illustrated in Figure~\ref{fig:vol-approximation}.

\subsection{Imitation Learning Problem Formulation}

\begin{problem}[Feedback Controller Imitation Learning]
\label{pb:learning}
Given a plant $\Plant$, a nominal  controller $\mathcal{C}$ such that the closed-loop system $\mathcal{C} || \Plant$ satisfies a PSTL specification $\Phi$, our problem is to learn a neural network controller $\mathcal{N}$ to imitate $\mathcal{C}$ such that
\begin{compactitem}
    \item the closed-loop system $ \mathcal{N} ||\Plant$
    satisfies $\Phi$, and
   \item the performance similarity $\similarity_{\Plant,\Phi}(\mathcal{C}, \mathcal{N})$ is as small as possible.
\end{compactitem}
\end{problem}
The learning guidance here is provided using positive examples, i.e. good behaviors, generated by the nominal controller which already satisfies the desired requirement. As we will see later, using parametric requirements allows more freedom in choosing nominal controllers satisfying some (minimal) performance, however to estimate the controller similarity in the learning process we estimate the actual Pareto-efficient parameters satisfied by a concrete nominal controller. Such a nominal controller may be complex and costly to execute and the ultimate goal is thus to use the learned NN controller to replace it. In the learning context, the nominal controller plays the role of a teacher that generates a desired control signal for a given system state which the NN should imitate. In this paper, we mostly focus on the problem of how to iteratively train NN using examples (good behaviors) and counter-examples (bad behaviors).




\section{Feedback Controller Learning Methodology}
\label{sec-methodology}

\def\NN{\text{NN}}

We will explain our approach for solving Problem~\ref{pb:learning} via an example that uses the PSTL formula $\Phi$ defined in~\eqref{eq:phi}. The major steps of our approach are as follows. Since initially we do not know the concrete performance of the nominal controller,  we can assume that we conservatively choose a parameter valuation $\pmax$ so that $\Phi(\pmax)$ is satisfied by the nominal controller. We train a NN controller satisfying $\Phi(\pmax)$ using an iterative neural network training approach which is detailed in the subsequent subsections. 

We generate and accumulate the traces generated by both controllers during the process. We use these traces to approximate their False domain. Then, we compute their volumes to estimate their policy similarity. Retraining is needed if this similarity is not as small as desired. To create data for such retraining, a finer grid can be used.


\subsection{Neural Network Structure and Training} Intuitively, the NN controller is trained based on the good closed-loop behaviors we want it to learn. Let $\traj = (x(\cdot), u(\cdot), y(\cdot))$ be a good closed-loop behavior, from which we extract the data of the form $\data_{\traj} = \bigl\{ (x_k, y_k, u_k) \bigm| k < K \bigr\}$
where $x_k$, $y_k$, and $u_k$ are respectively the state, output, and control values at time $t_k$; $K$ is the discrete-time horizon (that is, the number of sampled time points). When many behaviors are considered, the data set $\dataset$ is the union of all $\data_{\traj}$. We generate a neural network $\neuralnet$ to fit the data set $\dataset$. The structure of the NN should capture the input-output relationships of the nominal controller $\mathcal{C}$. We keep some past values to represent the memory needed to compute the output at each discrete step. The input of the NN is $(y_{k-1}, \ldots, y_{k-m_y}, u_{k-1}, \ldots, u_{k-m_u})$ where $n_z=m_y+m_u$ represents the dimension of the state variable of the controller. The NN output is $u_k$.  
To train the NN, we use a loss function  defined via the Root Mean Square Error (RMSE): $\sqrt{\displaystyle {\frac{1}{n_d}\sum_{k=1}^{n_d} (\|u_k-\overline{u}_k\|)^2}}$ where $u$ is the output of the nominal controller and $\overline{u}$ is the output of the NN. It is possible to use other loss functions such as MSE (Mean Square Error); in our experiments so far the RMSE metric is more convenient in terms of interpreting the effects of control input error. A data point is a pair of input and output values, the total number $n_d$ of data points is the number of data points per system behavior multiplied by the number of behaviors. 
The NN accuracy is defined by the {\em training} and {\em validation errors}, which are obtained from evaluating the loss function on the training and validation data.   

%



\subsection{Coverage based Data Generation}\label{sec:covData}


In order to achieve a robust NN controller, we need to provide data representing diverse settings that the NN should learn to cope with. Note that the set of reachable states of the system is infinite, we thus use a coverage measure based on $\varepsilon$-net~\cite{kolmogorov1959varepsilon} to quantify how well a finite set of sampled states covers the reachable set. 


We propose a simple grid-based method to construct $\varepsilon$-nets satisfying a separation requirement. These notions of $\varepsilon$-net and $\varepsilon$-separated sets are important to measure function approximation quality, expressed roughly as how to obtain an accurate function approximation with a small number of function evaluations.

Let us assume that the state space is a box $B_r = [\underline{r}, \overline{r}]^{n}$. We use a grid $\mathcal{G}$ to partition $B_r$ into a set $G$ of rectangular cells with equal side length $2\varepsilon$, assuming for simplicity that $(\overline{r} - \underline{r})/(2\varepsilon)$ is an integer. 
The set $C$ of center points of all the cells in $G$ is an $\varepsilon$-net of the state space. It can be proved that it is an $\varepsilon$-net with minimal cardinality and additionally a $\varepsilon$-separated set with maximal cardinality\footnote{This cardinality determines the $\varepsilon$-entropy and $\varepsilon$-capacity of the state space~\cite{kolmogorov1959varepsilon}.}. 



Datasets constructed using $\varepsilon$-nets guarantee $\varepsilon$ coverage. However, as we shall see later, the falsification procedure (to check if a neural network  satisfies the requirement) uses optimization-based search algorithms which can explore many states outside the $\varepsilon$-nets. To estimate a coverage measure that better reflects the portion of the tested behaviors, we can use a finer grid $\mathcal{G}_c$ and take the ratio between the number of cells visited by both the sampling and falsification procedures and the total number of cells in this grid $\mathcal{G}_c$.   


\section{Dataset Aggregation-based Training}
\label{sec-aggregation}

The top-level iterative training algorithm implements a data aggregation method, inspired by~\cite{ross2011reduction}, which uses the current NN to generate new data (from examples and counter-examples) that will be aggregated to the whole dataset to train a new NN. This dataset aggregation method considers the coverage and separation measures discussed in the previous section, to enable the NN to generalize well and provide high confidence in the correctness of the final NN. 

The dataset aggregation algorithm (Algorithm~\ref{algo:dagger}) determines whether it should generate new data for retraining or stop. It stops after a maximum number of iterations $k_{max}$ or as soon as the procedure responsible for providing new training samples fails to do so. 

\begin{algorithm}[bt!]
\caption{Dataset aggregation-based training algorithm.}
\label{algo:dagger}
\small
\begin{algorithmic}[1]
\State $\neuralnet_0 \gets \emptyset$, $\dataset_0 \gets \emptyset$, $k \gets 1$
\Repeat
    \State ($\dataset_k$, $\Status$) $\gets$ $\mathtt{getNewData}$\;($\neuralnet_{k-1}, \dataset_{k-1}$)
    \If{$\dataset_k \neq \dataset_{k-1}$} \Comment{In this case $\dataset_{k-1} \subset \dataset_{k}$}
        \State $\neuralnet_k \gets$ $\mathtt{Train}$\;($\dataset_k$) 
    \EndIf
    \State $k \gets k+1$
\Until{$\dataset_k = \dataset_{k-1}$ or $k>k_{max}$}
\State \Return{$\neuralnet_k$, $\Status$}
\end{algorithmic}
\end{algorithm}

The training procedure is successful if it terminates with $\dataset_k = \dataset_{k-1}$, {\em i.e.}, no new data is found, and the $\Status$ returned by $\mathtt{GetNewData}$ is ``No counter-example found''. This $\mathtt{GetNewData}$ procedure is described in Algorithm~\ref{algo:get_new_data}. 

\begin{algorithm}
  \caption{New data acquisition procedure.}   \label{algo:get_new_data}
\small
  \begin{algorithmic}[1]    
  \Procedure{getNewData}{$\dataset$, $\neuralnet$}
  \If{$\dataset$ is $\emptyset$} \Comment{Sample traces from initial set}
    \State $\InitSamples$ $\gets$ $\mathtt{getInitSamples}$\,(\,) 
    \State $\InitTraces$ $\gets$ $\mathtt{simNominal}$\,($\InitSamples$)
    \State $\dataset_{\text{new}}\gets$ $\mathtt{gridFilter}$\,($\InitTraces$)
    \State $\Status$ $\gets$ ``New data available for training."
  \Else \Comment{Search for counter-examples}
    \State $\CexTraces$ $\gets$ $\mathtt{falsify}$\,($\neuralnet$) 
    \If{$\CexTraces \neq\emptyset$} 
        \State ($\dataset_\text{new}$, $\Status$) $\gets$ $\mathtt{fixAndMerge}$\,($\dataset$, $\CexTraces$)         
    \Else \Comment{Success.}
        \State $\dataset_\text{new}\gets \dataset$
        \State $\Status$ $\gets$ "No counter-example found."        
    \EndIf
  \EndIf 
  \EndProcedure
\end{algorithmic}    
\end{algorithm}

\begin{figure}
    \centering
    \includegraphics[width=.45\textwidth, trim=10 5 10 10, clip]{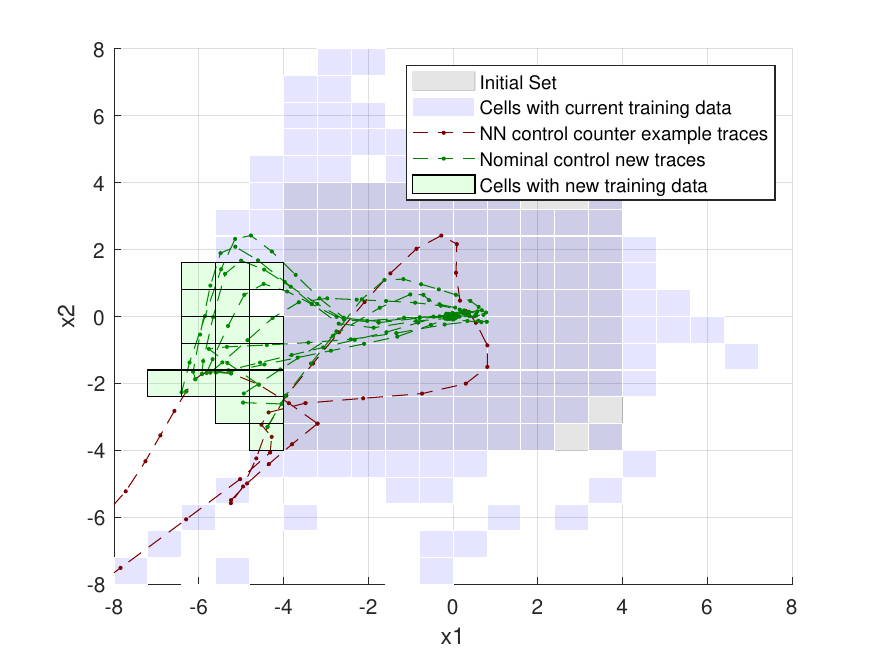}
    \caption{One iteration of the learning algorithm.}
    \label{fig:OneIteration}
\end{figure}

In the first call of Algorithm~\ref{algo:get_new_data}, when no neural network has been trained yet and the data set is empty, it samples states using an $\varepsilon$-net over the initial set via the $\mathtt{getInitSamples}$ function and computes nominal traces from these initial states, that is traces generated by the nominal controller via the $\mathtt{simNominal}$ function. The $\mathtt{gridFilter}$ function collects samples from these traces and ensures that only one sample per grid cell is kept so that the data remains $\varepsilon$-separated. Indeed the $\mathtt{gridFilter}$ function ``filters'' the sampled states by removing the newly sampled states from the cells that are covered by existing data already. A subset of the remaining samples is then used to compute new nominal traces, producing new data. This step is illustrated in Fig.~\ref{fig:OneIteration}.
At the subsequent calls, new data is obtained by testing the current neural network $\neuralnet_k$, looking for traces that violate the requirement using falsification ($\mathtt{falsify}$ function). The procedure $\mathtt{fixAndMerge}$ described in Algorithm~\ref{algo:fix_merge} is called to produce new nominal traces and add data from these new nominal traces, in order to ``fix'' these counter-examples. Again, to keep samples well separated, new nominal traces are generated only from the samples resulting after applying $\mathtt{gridFilter}$ on the bad traces ($\CexTraces$ - Line 2). We also remove samples that are already covered by $\dataset$ ($Cover(\dataset)$), represented as blue cells in Fig.~\ref{fig:OneIteration}).

Furthermore, in $\mathtt{selectNeigh}$ the samples close to the cells which are already covered have higher priority to be selected, which helps the neural net to generalize more easily. This can be seen in Fig.~\ref{fig:OneIteration} where the green cells (containing new data selected for retraining) are close to the blue cells (containing current training data).

\begin{algorithm}[t]
  \caption{Augmenting the training data from  bad traces.} \label{algo:fix_merge}
\small
  \begin{algorithmic}[1]  
  \Procedure{fixAndMerge}{$\dataset$, $\CexTraces$}
  \State $\NSamples$ $\gets$ $\mathtt{gridFilter}$\;($\CexTraces$)$\setminus \Cover(\dataset)$  
  \If{$\NSamples$ $=\emptyset$} 
    \State $\Status$ $\gets$ ``Counter-examples do not add new data.'' 
    \State $\dataset_\text{new}\gets \dataset$
  \Else \Comment{Get fixed traces from some bad samples}
    \State $\NSamples$ $\gets$ $\mathtt{selectNeigh}$\;($\NSamples$)   
    \State $\FixedTraces$ $\gets$ $\mathtt{simNominal}$\;($\NSamples$)
    \State $\dataset_{\text{new}}\gets$ $\mathtt{gridFilter}$\;($\dataset\ \cup$ $\FixedTraces$)
    \State $\Status$ $\gets$``New data available for training.''
  \EndIf  
  \EndProcedure
\end{algorithmic}    
\end{algorithm}

Algorithm~\ref{algo:dagger} stops either when no new data is returned by Algorithm~\ref{algo:get_new_data} or when $k_{max}$ is reached. In the first case (no new data), there are two situations:
\begin{compactitem}
    \item ``No counter-example found'': Successful case. The falsification procedure has failed to falsify the neural network controller, providing our strongest evidence that we have obtained a correct controller with test coverage at least $\varepsilon$. 
    \item ``Counter-examples do not add new data.''. This means that the grid is likely too coarse and the counter-examples do not allow visiting new cells. 
\end{compactitem}

If the maximum number of iterations is reached, the two statuses above are still possible (both stopping conditions are met), otherwise, the status returned is: ``New data available for training.'' which indicates that the user can resume the process using the data aggregated so far.

\section{Flying Robot Case Study}
\label{sec-casestudy}
This example is taken from Mathworks \cite{flying_robot_mathworks} and describes a model of a flying robot that is driven by a nonlinear model predictive controller. The flying robot has two thrusters to move it in a 2-D space. The state of the flying robot, denoted by $x$, consists of six components:     $x_1$, 
 $x_2$ (horizontal and vertical coordinate) $\theta$  (robot thrust direction), and their derivatives $\dot{x}_1$, $\dot{x}_2$ and $\dot{\theta}$. The thrusts are represented by $u = (u_1, u_2)$. The dynamics of the robot are given by $\ddot{x_1}=(u_1+u_2)\cdot \cos(\theta)$, $\ddot{x_2}=(u_1+u_2)\cdot \sin(\theta)$ and 
$\ddot{\theta}= \alpha \cdot u_1+\beta\cdot u_2$. The parameters are $\alpha=0.2$ and $\beta=0.2$. For each thrust, there is an operating range $[-u_{\mathit{max}}, u_{\mathit{max}}]$ with $u_{\mathit{max}}=3$. Considering the following set $X_0$ of initial conditions 
$$
  \begin{array}{rl}
 X_0=\left\{\right.& \left(x_1(0),x_2(0),\theta(0), \dot{x}_1(0),\dot{x}_2(0),\dot{\theta}(0)\right)\ s.t. \\    
    &x_1(0),x_2(0)\in [3.8, 3.8],\theta(0)\in [-2.5, 2.5],\\
    &\left. \dot{x}_1(0),\dot{x}_2(0)\in[-1.6,1.6],\dot{\theta}(0) \in [-0.8,0.8]\right\},
\end{array}
$$
the goal of the control input is to drive the flying robot from any state in $X_0$ to a region close to the origin and maintain it in this region indefinitely. We define $y$ as $y(t)=\|x(t)\|$, so that the problem becomes to stabilize $y$ close to 0. The nominal controller is provided by a model predictive control (MPC) scheme which computes inputs at each step that minimizes $y(t)$ over a given horizon. The simulation time is 15 seconds. To evaluate the performance of a controller for this problem, we used the PSTL formula $\Phi$ defined in Equation~\ref{eq:phi}, where we set $\taustab$ to $+\infty$ and  considered overshoot $\piov$, transient time $\tautrans$ and stabilization region $\pistab$ as variables to measure. We applied our algorithm with the STL formula $\Phi(\pmax)$ with $\pmax= \{ \piov \rightarrow 15, \tautrans \rightarrow 14, \pistab \rightarrow 2\}$.

\begin{figure}
    \centering
    \includegraphics[width=.5\textwidth, trim=0 10 0 188, clip]{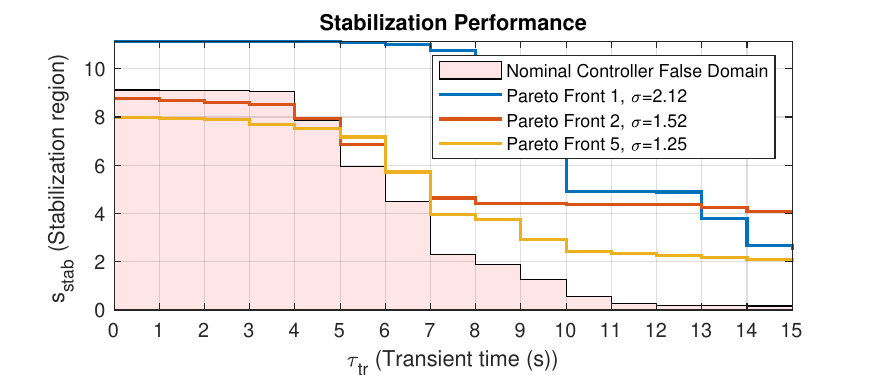}
    \caption{Result of training NN controllers for the flying robot. Good performance is obtained after only 5 iterations. The red region (Nominal Controller False Domain) represents the valuations $p$ for which $\Phi(p)$ are not satisfied by the nominal controller. The boundary of this region represents the Pareto front of the nominal controller. Other plots represent the Pareto fronts for several instances of the NN controllers computed for different iterations. Similarity with the nominal MPC controller is indicated in the legend.} 
    \label{fig:pareto}
\end{figure}

The inputs of NNs we trained correspond to the state variables of the system and the previous control input variables. They have RELU as activation functions, $6$ hidden layers (each with $256$ neurons), one scaling layer and one output layer. Figure~\ref{fig:pareto} shows the performance evaluations of the nominal controller and a sample of different trained NN controllers obtained after 6 iterations, computed in around 120 minutes of computation time on a standard PC with an Intel Core i7 10700 processor and 64 Go of memory.  After only two iterations, the NN controller manages to improve on the overshoot. However, looking at the False domain for  ($\taustab, \tautrans$), we can see that the stabilization region is larger than that of the MPC controller, meaning that neural networks have trouble stabilizing the robot close to the origin due to the inherent instability of the system at this state.
  
\section{Conclusion}

In this paper, we have presented a framework for efficiently training a neural network-based controller by imitation learning using a dataset aggregation approach with several novel aspects. Most notably, the collection of data from the nominal (expert) controller is done at states where the trained controller caused the plant to fail according to a specification expressed in Signal Temporal Logic. Moreover, the same specification is parameterized and can be used to evaluate the performance of the trained controller and how far or how differently it behaves with respect to the nominal controller. The method was evaluated on a nonlinear robotic system with promising results. Further experiments will be conducted and different practical and theoretical questions remain to be explored but we believe that this work represents an interesting step in the direction of safer and more efficient imitation learning methods of complex control systems. 


\bibliographystyle{IEEEtran}
\bibliography{biblio,learning_control,nn_verification,formal}

\end{document}